%% file: main.tex
\documentclass[10pt,twocolumn,letterpaper]{article}

\usepackage{cvpr}              

\input{header}

\definecolor{cvprblue}{rgb}{0.21,0.49,0.74}
\usepackage[pagebackref,breaklinks,colorlinks,citecolor=cvprblue]{hyperref}

\def\paperID{14} 
\def\confName{CVPR}
\def\confYear{2024}

\title{Posterior Distillation Sampling}

\author{Juil Koo $\quad$
Chanho Park $\quad$
Minhyuk Sung \\
KAIST \\
{\tt\small \{63days,charlieppark,mhsung\}@kaist.ac.kr}
}

\begin{document}
\twocolumn[{%
\renewcommand\twocolumn[1][]{#1}%
\maketitle
\vspace{-30pt}
\begin{center}
\centering
\captionsetup{type=figure}
\includegraphics[width=\textwidth]{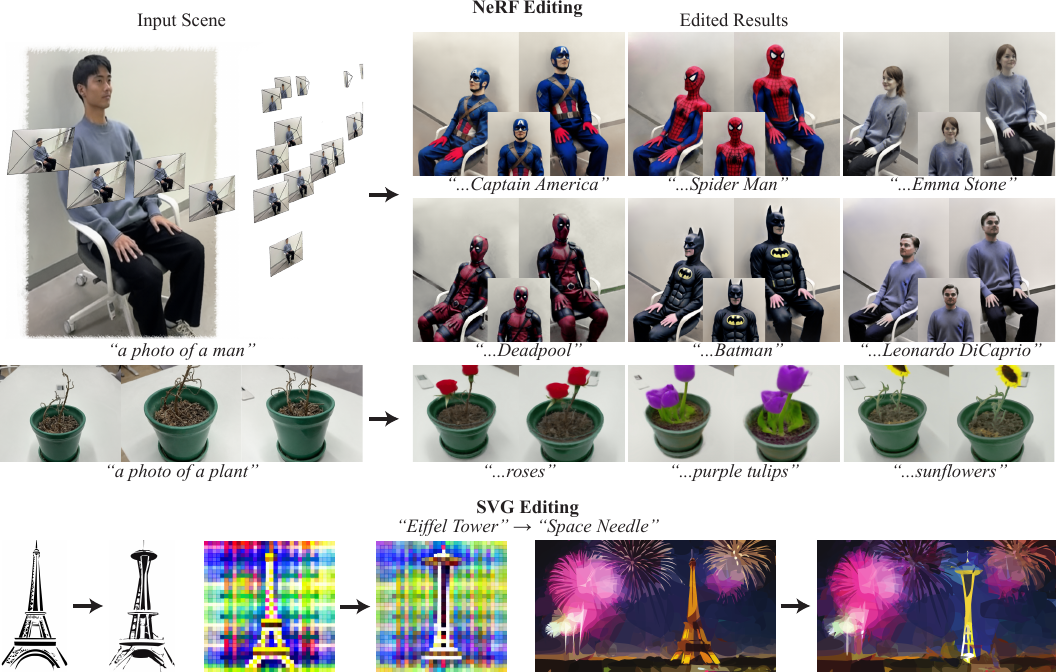}
\vspace{-10pt}
\caption{\textbf{Parametric image editing results obtained by Posterior Distillation Sampling (PDS).} PDS is an optimization tailored for editing across diverse parameter spaces. It preserves the original details of the source content while aligning them with the input texts.}

\label{fig:teaser}
\end{center}
}]

\newif\ifpaper
\papertrue

\input{Sections/0_Abstract}    
\input{Sections/1_Introduction}
\input{Sections/2_Related_Work}
\input{Sections/3_Background}
\input{Sections/4_Method}
\input{Sections/5_Experimental_Results}

\input{Sections/6_Conclusion}

{
    \small
    \bibliographystyle{ieeenat_fullname}
    \bibliography{main}
}

\renewcommand{\thesection}{A}
\renewcommand{\thetable}{A\arabic{table}}
\renewcommand{\thefigure}{A\arabic{figure}}

\ifpaper
\clearpage
\newpage
\onecolumn
\section*{Appendix}
\input{Sections/999_Supplementary}
\else
\fi 
\end{document}

%% file: header.tex
\usepackage[dvipsnames]{xcolor}

\usepackage{amsmath}
\usepackage{amssymb}
\usepackage{booktabs}
\usepackage{dsfont}
\usepackage{graphicx}
\usepackage{makecell}
\usepackage{multirow}
\usepackage{pifont}
\usepackage[group-separator={,}]{siunitx}
\usepackage{tabularx}
\usepackage{xcolor}
\usepackage{caption}
\usepackage{soul}
\usepackage{changepage,threeparttable}
\usepackage{colortbl}
\usepackage[accsupp]{axessibility}  
\usepackage{mathtools}
\usepackage{xfrac}
\usepackage{catchfile}
\usepackage{longtable}
\usepackage{tabu}
\usepackage{tcolorbox}
\usepackage{supertabular}
\usepackage{algpseudocode}
\usepackage{algorithm}
\newcolumntype{Y}{>{\centering\arraybackslash}X}
\usepackage{physics}

\renewcommand{\ie}{i.e.,}
\renewcommand{\eg}{e.g.,}
\renewcommand{\etal}{\emph{et al.\ }}

\definecolor{color_1}{RGB}{255,0,128}
\definecolor{color_2}{RGB}{0,128,128}
\definecolor{color_3}{RGB}{0,128,0}
\definecolor{color_4}{RGB}{128,0,0}
\definecolor{color_5}{RGB}{128,0,128}
\definecolor{cadetgrey}{RGB}{0.57, 0.64, 0.69}

\newcommand{\Veps}{\boldsymbol{\epsilon}}
\newcommand{\Vmu}{\boldsymbol{\mu}}

\newcommand{\B}{\mathbf}
\newcommand{\Src}{\text{src}}
\newcommand{\Tgt}{\text{tgt}}
\newcommand{\Ours}{PDS}

%% file: Sections/0_Abstract.tex
\begin{abstract}
\ifpaper
\vspace{-1.0\baselineskip}
\else
\vspace{-1.3\baselineskip}
\fi 
We introduce Posterior Distillation Sampling (PDS), a novel optimization method for parametric image editing based on diffusion models. 
Existing optimization-based methods, which leverage the powerful 2D prior of diffusion models to handle various parametric images, have mainly focused on generation. Unlike generation, editing requires a balance between conforming to the target attribute and preserving the identity of the source content. Recent 2D image editing methods have achieved this balance by leveraging the stochastic latent encoded in the generative process of diffusion models. To extend the editing capabilities of diffusion models shown in pixel space to parameter space, we reformulate the 2D image editing method into an optimization form named PDS. PDS matches the stochastic latents of the source and the target, enabling the sampling of targets in diverse parameter spaces that align with a desired attribute while maintaining the source's identity. We demonstrate that this optimization resembles running a generative process with the target attribute, but aligning this process with the trajectory of the source's generative process. Extensive editing results in Neural Radiance Fields and Scalable Vector Graphics representations demonstrate that PDS is capable of sampling targets to fulfill the aforementioned balance across various parameter spaces. 
\ifpaper
Our project page is at \href{https://posterior-distillation-sampling.github.io/}{https://posterior-distillation-sampling.github.io}.
\else 
\fi
\end{abstract}
\vspace{-\baselineskip}

%% file: Sections/1_Introduction.tex
\begin{figure*}
\centering
\includegraphics[width=\textwidth]{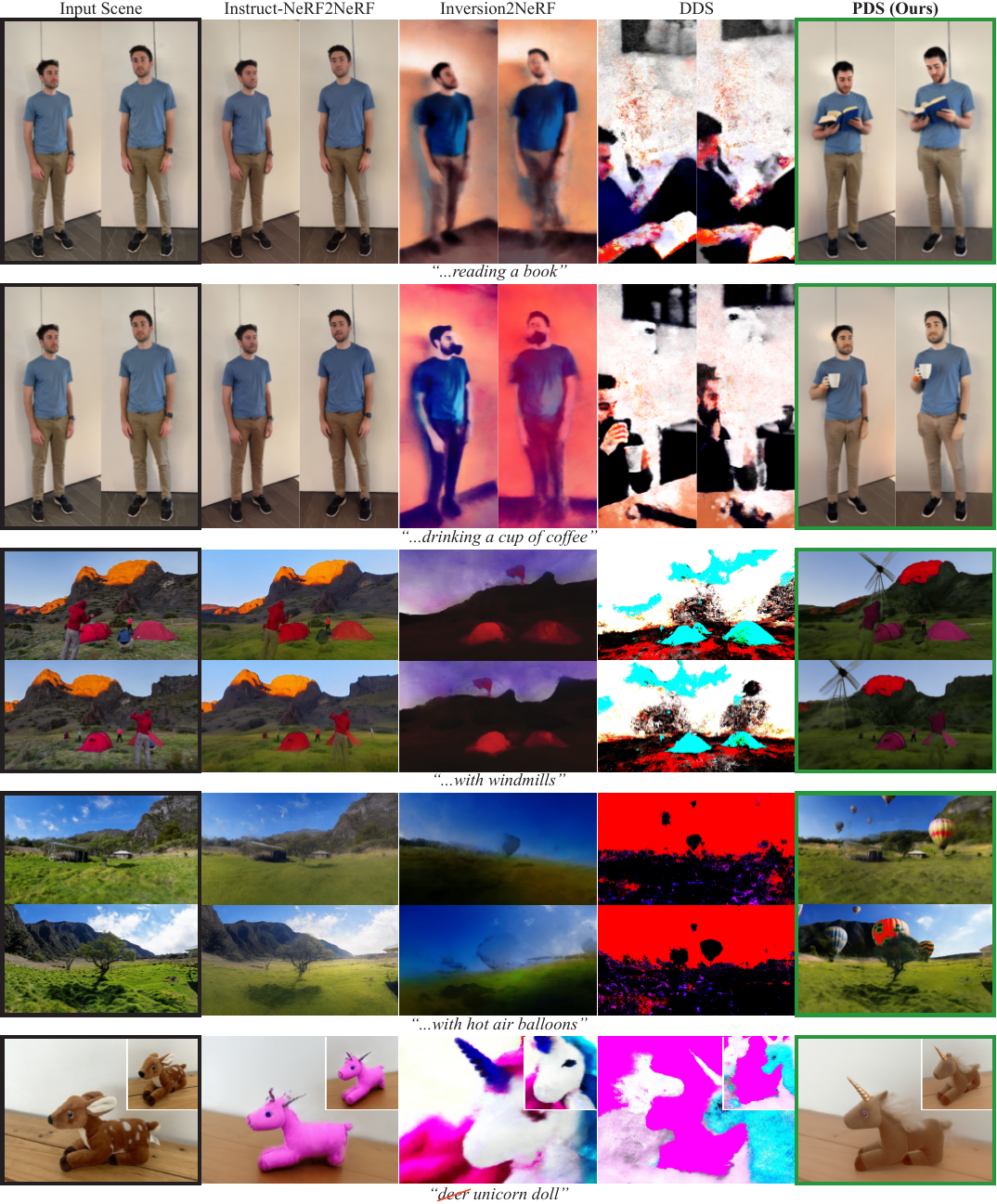}
\caption{\textbf{A comparison of 3D scene editing between \Ours{} and other baselines.} Given input 3D scenes on the left, \Ours{}, marked by \textcolor{Green}{green boxes} on the rightmost side, successfully performs complex editing, such as geometric changes and adding objects, according to the input texts. On the other hand, the baselines either fail to change the input 3D scenes or produce results that greatly deviate from the input scenes, losing their identity.}
\label{fig:nerf_editing_comparison}
\end{figure*}

\section{Introduction}
\label{sec:introduction}
Diffusion models~\cite{Ho:2020DDPM, Song:2021DDIM, Song:2021SGM, Sohl-Dickstein:2015Deep, Song:2019NCSN} have recently led to rapid development in text-conditioned generation and editing across diverse domains, including 2D images~\cite{Lee:2023SyncDiffusion, Wallace:2023EDICT, Huberman:2023FriendlyInversion, Wu:2023CycleDiffusion, Hertz:2023Prompt-to-Prompt}, 3D objects~\cite{Jun:2023ShapE, Nichol:2022Point-E, Li:2023Diffusion-SDF, Koo:2023SALAD}, and audio~\cite{Huang:2023Make-an-Audio, Ghosal:2023Tango, Yang:2023DiffSound}. Among them, in particular, 2D image diffusion models~\cite{Ramesh:2022DALLE2, Rombach:2022StableDiffusion, Saharia:2022IMAGEN, DeepFloyd, Midjourney} have demonstrated their powerful generative prior aided by Internet-scale image and text datasets~\cite{Schuhmann:2022LAION5B, Schuhmann:2021LAION400M, Kakaobrain:2022COYO-700M}. Nonetheless, this rich 2D generative prior has been confined to pixel space, limiting their broader applicability. A pioneer work overcoming this limitation, DreamFusion~\cite{Poole:2023DreamFusion}, has introduced Score Distillation Sampling (SDS). It leverages the generative prior of text-to-image diffusion models to synthesize 3D scenes represented by Neural Radiance Fields (NeRFs)~\cite{Mildenhall:2020NeRF} from texts. Beyond NeRF representations~\cite{Lin:2023Magic3D, Wang:2023ProlificDreamer, Shi:2023MVDream, Zhu:2023HiFA, raj2023dreambooth3d, Chen:2023Fantasia3D, Wang:2023SJC}, SDS has been widely applied to various parameter spaces, where images are not represented by pixels but specific parameterizations, such as texture~\cite{Metzer:2023LatentNeRF, Anonymous:2023SDSTexturing}, material~\cite{Xu:2023MATLABER} and Scalable Vector Graphics (SVGs)~\cite{Jain:2023VectorFusion, Xing:2023DiffSketcher, Iluz:r2023Word-As-Image}. 

While SDS~\cite{Poole:2023DreamFusion} has achieved great advances in generating parametric images, editing is also an essential element for full freedom in handling visual content. Editing differs from generation in that it requires considerations of both the target text and the original source content, thereby emphasizing two key aspects: (1) alignment with the target text prompt and (2) preservation of the source content's identity. To extend SDS, which lacks the latter aspect,  Hertz~\etal~\cite{Hertz:2023DDS} propose Delta Denoising Score (DDS). DDS reduces the noisy gradients inherent in SDS, leading to better-maintaining background details and sharper editing outputs. However, the optimization function of DDS still lacks an explicit term for identity preservation.

To address the absence of preserving the source's identity in SDS~\cite{Poole:2023DreamFusion} and DDS~\cite{Hertz:2023DDS}, we turn our attention to a recent 2D image editing method~\cite{Wu:2023CycleDiffusion, Huberman:2023FriendlyInversion} based on diffusion models, known as stochastic diffusion inversion. Their primary objective is to compute the stochastic latent of an input image within the generative process of diffusion models. Once the stochastic latent of a source image is computed, the source image can be edited by running a generative process with new conditions, such as new target text prompts, while feeding the source's stochastic latent into the process. Feeding the source's stochastic latent into the target image's generative process ensures that the target image maintains the structural details of the source while moving towards the direction of the target text. Thus, this editing process reflects the aforementioned two key aspects of editing.

To extend the editing capabilities of the stochastic diffusion inversion method from pixel space to parameter space, we reformulate this method into an optimization form named Posterior Distillation Sampling (PDS). Unlike SDS~\cite{Poole:2023DreamFusion} and DDS~\cite{Hertz:2023DDS}, which match two noise variables, PDS aims to match the stochastic latents of the source and the optimized target. We demonstrate that our optimization process resembles aligning forward process posteriors of the source and the target, ensuring that the target's generative process trajectory does not significantly deviate from that of the source.  

When parametric images come from NeRF~\cite{Mildenhall:2020NeRF}, Haque~\etal~\cite{Haque:2023InstructNeRF} have recently introduced a promising text-driven NeRF editing method called Iterative Dataset Update (Iterative DU). To edit 3D scenes, it performs an editing process in 2D space bypassing direct edit in 3D space. Thus, when a text prompt induces large variations in 2D space across different views, it has difficulty producing the right edit in 3D space. On the other hand, our method directly updates NeRF in 3D space, thus gradually transforming a 3D scene into its edited version in a view-consistent manner even in the case where text prompts induce large variations, such as large geometric changes or the addition of objects to unspecified regions.

Our extensive editing experiment results, including NeRF editing (Section~\ref{sec:nerf_editing}) and SVG editing (Section~\ref{sec:svg_editing}), demonstrate the versatility of our method for parametric image editing. In NeRF editing, we are the first to produce large geometric changes or to add objects to arbitrary regions without specifying local regions to be edited. Figure~\ref{fig:nerf_editing_comparison} shows these examples. Qualitative and quantitative comparisons of SVG editing with other optimization methods, namely SDS~\cite{Poole:2023DreamFusion} and DDS~\cite{Hertz:2023DDS}, have demonstrated that \Ours{} produces only the necessary changes to source SVGs, effectively aligning them with the target prompts. 

%% file: Sections/2_Related_Work.tex
\section{Related Work}
\label{secl:related_work}
\subsection{Score Distillation Sampling}
Following the remarkable success of diffusion models in text-to-image generation, there have been attempts to leverage the 2D prior of diffusion models for various other types of generative tasks. In these tasks, images are represented through rendering processes with specific parameters, including Neural Radiance Fields~\cite{Poole:2023DreamFusion, Wang:2023SJC, Jain:2023VectorFusion}, texture~\cite{Anonymous:2023SDSTexturing, Metzer:2023LatentNeRF}, material~\cite{Xu:2023MATLABER} and Scalable Vector Graphics (SVGs)~\cite{Jain:2023VectorFusion, Xing:2023DiffSketcher, Iluz:r2023Word-As-Image}. The primary method employed in these tasks is Score Distillation Sampling (SDS).
SDS is an optimization approach that updates the rendering parameter towards the image distribution of diffusion models by enforcing the noise prediction on noisy rendered images to match sampled noise. Concurrently, Wang~\etal~\cite{Wang:2023SJC} also have introduced Score Jacobian Chaining which converges toward a similar algorithm as SDS but from a different mathematical derivation. Wang~\etal~\cite{Wang:2023ProlificDreamer} have proposed Variational Score Distillation (VSD) to address over-saturation, over-smoothing, and low-diversity problems in SDS~\cite{Poole:2023DreamFusion}. Instead of updating a single data point, VSD updates multiple data points to align an optimized distribution with the diffusion model's image distribution. Zhu and Zhuang~\cite{Zhu:2023HiFA} use more accurate predictions of diffusion models via iterative denoising at every SDS update step. 

When it comes to editing, Hertz~\etal~\cite{Hertz:2023DDS} propose Delta Denoising Score (DDS), an adaptation of SDS for editing tasks. It reduces the noisy gradient directions in SDS to better maintain the input image details. Nonetheless, its optimization function lacks an explicit term to preserve the identity of the input image, thus often producing outputs that significantly deviate from the input images. To alleviate this issue, we propose Posterior Distillation Sampling, a novel optimization approach that incorporates a term dedicated to preserving the identity of the source in its optimization function.

\subsection{Text-Driven NeRF Editing}
Haque~\etal~\cite{Haque:2023InstructNeRF} have proposed a text-driven NeRF editing method, known as Iterative Dataset Update (Iterative DU). It iteratively replaces reference images, initially used for NeRF~\cite{Mildenhall:2020NeRF} reconstruction, with edited images using Instruct-Pix2Pix~\cite{Brooks:2023InstructPix2Pix}. By applying a reconstruction loss with these iteratively updated images to an input NeRF~\cite{Mildenhall:2020NeRF} scene, the scene is gradually transformed to its edited counterpart.
Mirzae~\etal~\cite{Mirzae:i2023WatchYourStep} improve Instruct-NeRF2NeRF~\cite{Haque:2023InstructNeRF} by computing local regions to be edited. However, this iterative image replacement method suffers from edits that involve large variations across different views, such as complex geometric changes or adding objects to unspecified regions. Thus, they have mainly focused on appearance changes. 

Instead of the Iterative DU method, several recent works~\cite{Park:2023ED-NeRF, Li:2023FocalDreamer, Zhuang:2023DreamEditor} directly apply SDS~\cite{Poole:2023DreamFusion} or DDS~\cite{Hertz:2023DDS} to NeRF editing. However, these optimizations do not fully consider the preservation of the source's identity and are thus prone to producing outputs that substantially diverge from the input scenes. In contrast, our novel optimization inherently guarantees the preservation of the source's identity, facilitating involved NeRF editing while maintaining the identity of the original scene.

\subsection{Diffusion Inversion}
Diffusion inversion computes the latent representation of an input image encoded in diffusion models. This allows for real image editing by finding the corresponding latent that can fairly reconstruct the given image. The computed latent is then decoded into a new image through a generative process. Using the deterministic generative process of Denoising Diffusion Implicit Models (DDIM)~\cite{Song:2021DDIM}, one can approximately run the ODE of the generative process in reverse~\cite{Song:2021DDIM, Dhariwal:2021ADM}, referred to as DDIM inversion. Several recent works have improved DDIM inversion by adjusting text features~\cite{Mokady:2023NullTextInversion, Han:2023ProximalGuidance, Miyake:2023NegativePrompt}, introducing new cross-attention maps during a generative process~\cite{Hertz:2023Prompt-to-Prompt} or alternatively coupling intermediate latents from two inversion trajectories~\cite{Wallace:2023EDICT}. Meanwhile, an alternative approach, known as DDPM inversion~\cite{Huberman:2023FriendlyInversion, Wu:2023CycleDiffusion}, employs the stochastic generative process of Denoising Diffusion Probabilistic Models (DDPM)~\cite{Ho:2020DDPM}. They focus on capturing the structural details of an input image encoded in its stochastic latent. We extend the editing capabilities of this DDPM inversion method to parameter space by reformulating the method into an optimization form.

%% file: Sections/3_Background.tex
\section{Preliminaries}
We first discuss existing optimization-based approaches to handle parametric images, then introduce our novel parametric image editing method in Section~\ref{sec:pds}.

\subsection{Score Distillation Sampling (SDS)~\cite{Poole:2023DreamFusion}}
Score Distillation Sampling (SDS)~\cite{Poole:2023DreamFusion} is proposed to generate parametric images by leveraging the 2D prior of pre-trained text-to-image diffusion models. Given an input data $\B{x}_0$ and a text prompt $y$, the training objective function of diffusion models is to predict injected noise $\Veps$ using a noise predictor $\Veps_\phi$:
\begin{align}
\label{eq:diffusion_training_objective}
    \mathcal{L}(\B{x}_0) = \mathbb{E}_{t\sim \mathcal{U}(0, 1), \Veps_t} \left[
    w(t)\Vert \Veps_\phi(\B{x}_t,y, t) - \Veps_t \Vert_2^2\right],
\end{align}
where $w(t)$ is a weighting function and $\B{x}_t$ results from the forward process of diffusion models:
\begin{align}
    \B{x}_t := \sqrt{\bar\alpha_t}\B{x}_0 + \sqrt{1 - \bar\alpha_t}\Veps_t,\quad \Veps_t \sim \mathcal{N}(\B{0}, \B{I}) 
\end{align}
with variance schedule variables $\bar{\alpha}_t := \prod_{s=1}^t \alpha_s$.
When the input data $\B{x}_0$ is generated by a differentiable image generator $\B{x}_0 = g(\theta)$, parameterized by $\theta$, SDS 
updates $\theta$ by backpropagating the gradient of Equation~\ref{eq:diffusion_training_objective} while omitting the U-Net jacobian term $\frac{\partial \Veps_\phi}{\partial \B{x}_t}$ for computation efficiency:
\begin{align}
\label{eq:SDS}
    \nabla_\theta \mathcal{L}_{\text{SDS}}(\B{x}_0=g(\theta))=
     \mathbb{E}_{t,\Veps_t} \left[ w(t)(\Veps_\phi(\B{x}_t,y,t) - \Veps_t)\frac{\partial \B{x}_0}{\partial \theta} \right],
\end{align}
where we denote a noise prediction of diffusion models with classifier-free guidance~\cite{Ho:2021CFG} by $\Veps_\phi$ for simplicity.
Through this optimization process, SDS is capable of generating a parametric image which conforms to the input text prompt $y$. 

\subsection{Delta Denoising Score (DDS)~\cite{Hertz:2023DDS}}
Even though SDS has been widely used for various parametric images, its optimization is designed for generation, thus it does not reflect one of the key aspects of editing: preserving the source identity.

To extend SDS to editing, Hertz~\etal~\cite{Hertz:2023DDS} have proposed Delta Denoising Score (DDS). Given source data $\B{x}^{\text{src}}$ and its corresponding text prompt $y^{\text{src}}$, the goal of DDS is to synthesize new target data $\B{x}^{\text{tgt}}$ that is aligned with a target text prompt $y^{\text{tgt}}$. In the SDS formula~\ref{eq:SDS}, DDS replaces randomly sampled noise $\Veps$ with a noise prediction given a source data-text pair $\Veps_\phi(\B{x}^{\text{src}}_t,y^{\text{src}},t)$:
\begin{align}
\label{eq:DDS}
    &\nabla_\theta\mathcal{L}_{\text{DDS}} = \nonumber \\ 
    &\mathbb{E}_{t,\Veps_t} \left[w(t) \left( \Veps_\phi(\B{x}^\Tgt_t,y^\Tgt,t) - \Veps_\phi(\B{x}^\Src_t,y^{\text{src}},t) \right) \frac{\partial\B{x}^\Tgt_0}{\partial \theta}\right],
\end{align}
where the same noise $\Veps_t$ is shared for $\B{x}_t^{\text{src}}$ and $\B{x}_t^{\text{tgt}}$:
\begin{align}
\label{eq:dds_share_same_noise}
    \Veps_t &\sim \mathcal{N}(\B{0}, \B{I}), \nonumber\\
    \B{x}_t^\Src &= \sqrt{\bar{\alpha}_t}\B{x}_0^\Src + \sqrt{1 - \bar{\alpha}_t} \Veps_t, \nonumber\\
    \B{x}_t^\Tgt &= \sqrt{\bar{\alpha}_t}\B{x}_0^\Tgt + \sqrt{1 - \bar{\alpha}_t} \Veps_t.
\end{align}

While DDS extends SDS for editing tasks, it lacks an explicit term in its optimization to preserve the identity of the source. As a result, DDS is still prone to produce editing results that significantly deviate from the source. 

\subsection{Stochastic Latent in Generative Process}
To achieve both conformity to the text and preservation of the source's identity, we turn our attention to the rich information encoded in the stochastic generative process of DDPM~\cite{Ho:2020DDPM}.
When $\beta_t := 1-\alpha_t$ are small, it is well-known that the posterior of the forward process also follows a Gaussian distribution according to a property of Gaussians. The forward process posteriors are represented as:
\begin{align}
\label{eq:forward_process_posterior}
    q(\B{x}_{t-1} | \B{x}_t, \B{x}_0) = \mathcal{N}(\Vmu(\B{x}_t, \B{x}_0), \sigma_t \B{I}),
\end{align}
where $\sigma_{t} := \frac{1 - \bar\alpha_{t-1}}{1 - \bar\alpha_{t}} \beta_t$ and the posterior mean $\Vmu$ is a linear combination of $\B{x}_0$ and $\B{x}_{t}$: $\Vmu(\B{x}_{t}, \B{x}_0) := \gamma_t\B{x}_0 + \delta_t\B{x}_{t}$ with $\gamma_t:=\frac{\sqrt{\bar\alpha_{t-1}}(1 -\alpha_{t})}{1 - \bar\alpha_{t}}$ and $\delta_t:=\frac{\sqrt\alpha_{t} (1 - \bar\alpha_{t-1})}{1 - \bar\alpha_{t}}$.

Since $\B{x}_0$ is unknown during a generative process, we approximate $\B{x}_0$ with a one-step denoised estimate as follows:  
\begin{align}
\label{eq:tweedie}
    \tilde{\B{x}}_0(\B{x}_{t}, y; \Veps_\phi) :=
    \frac{1}{\sqrt{\bar\alpha_{t}}}(\B{x}_{t} - \sqrt{1 - \bar\alpha_{t}}\Veps_\phi(\B{x}_{t}, y, t)).
\end{align}
Consequently, one step of the generative process is represented as follows:
\begin{align}
\label{eq:one_step_generative_process}
    \B{x}_{t-1} = \Vmu_\phi(\B{x}_t, y;\Veps_\phi) + \sigma_t \B{z}_t, \quad \B{z}_t\sim\mathcal{N}(\B{0}, \B{I}),
\end{align}
where $\Vmu_\phi(\B{x}_t, y;\Veps_\phi) = \gamma_t \tilde{\B{x}}_0(\B{x}_t, y; \Veps_\phi) + \delta_t \B{x}_t$.

Using Equation~\ref{eq:one_step_generative_process}, one can compute stochastic latent $\tilde{\B{z}}_t$ that captures the structural details of $\B{x}_0$. This involves computing $\B{x}_t$ and $\B{x}_{t-1}$ via the forward process and then rearranging Equation~\ref{eq:one_step_generative_process} as follows:
\begin{align}
    \tilde{\B{z}}_t(\B{x}_0,y;\Veps_\phi) = \frac{\B{x}_{t-1} - \Vmu_\phi(\B{x}_t,y ; \Veps_\phi)}{\sigma_t}.
\end{align}

Several recent works~\cite{Wu:2023CycleDiffusion, Huberman:2023FriendlyInversion}, known as DDPM inversion, have utilized the stochastic latent for image editing tasks. To edit an image using $\tilde{\B{z}}_t$, they first pre-compute $\tilde{\B{z}}_t$ of the source image across all $t$ in the generative process. They then run a new generative process with a new target prompt while incorporating the pre-computed $\tilde{\B{z}}_t$ of the source into the process instead of randomly sampled noise $\B{z}_t$.

Although these works~\cite{Wu:2023CycleDiffusion, Huberman:2023FriendlyInversion} have utilized the rich information encoded in $\tilde{\B{z}}_t$ for an editing purpose, their applications have been limited within 2D-pixel space due to reliance on the generative process. In our work, we broaden the application of the stochastic latent to parameter space by reformulating the method as an optimization form, enabling \emph{parametric image editing}.

%% file: Sections/4_Method.tex
\section{Posterior Distillation Sampling}
\label{sec:pds}
Here, we introduce Posterior Distillation Sampling (PDS), a novel optimization function designed for parametric image editing. 

Our objective is to synthesize $\B{x}_0^\Tgt$ that is aligned with $y^\Tgt$ while it retains the identity of $\B{x}_0^\Src$. To achieve this, we employ the stochastic latent $\tilde{\B{z}}_t$ in our optimization.
For simplicity, we denote the stochastic latents of the source and the target as follows:
\begin{align}
    \tilde{\B{z}}_{t}^{\text{src}}&:=\tilde{\B{z}}_t(\B{x}_0^\Src, y^\Src; \Veps_\phi) \\
    \tilde{\B{z}}_{t}^{\text{tgt}}&:=\tilde{\B{z}}_t(\B{x}_0^\Tgt, y^\Tgt; \Veps_\phi).
\end{align}
Using the stochastic latents, we define a novel objective function as follows:
\begin{align}
    \mathcal{L}_{\tilde{\B{z}}_t}(\B{x}_0^{\text{tgt}} = g(\theta)) :=
    \mathbb{E}_{t, \Veps_{t-1}, \Veps_{t}}\left[
    \Vert \tilde{\B{z}}_{t}^{\text{tgt}} - \tilde{\B{z}}_{t}^{\text{src}} \Vert_2^2\right],
\end{align}
where, similar to Equation~\ref{eq:dds_share_same_noise}, $\tilde{\B{z}}^\Src_t$ and $\tilde{\B{z}}^\Tgt_t$ share the same noises, denoted by $\Veps_{t-1}$ and $\Veps_{t}$, when computing their respective $\B{x}_{t-1}$ and $\B{x}_t$.

Rather than matching noise variables as in SDS~\cite{Poole:2023DreamFusion} and DDS~\cite{Hertz:2023DDS}, we match the stochastic latents of the source and the target via the optimization. 
By taking the gradient of $\mathcal{L}_{\tilde{\B{z}}_t}$ with respect to $\theta$ and ignoring the U-Net jacobian term as previous works~\cite{Poole:2023DreamFusion, Hertz:2023DDS, Wang:2023SJC}, one can obtain \Ours{} as follows:
\begin{align}
\label{eq:DIS}
    \nabla_\theta \mathcal{L}_{\text{\Ours{}}}:=
    \mathbb{E}_{t, \Veps_t, \Veps_{t-1}}\left[
    w(t)(\tilde{\B{z}}_{t}^\Tgt - \tilde{\B{z}}_{t}^\Src) \frac{\partial\B{x}_0^\Tgt}{\partial \theta}\right].
\end{align}
Expanding Equation~\ref{eq:DIS}, the following detailed formulation is derived:
\begin{align}
\label{eq:final_pds_form}
    &\nabla_\theta \mathcal{L}_{\text{\Ours{}}} :=  \nonumber \\
   &\mathbb{E}_{t, \Veps_t, \Veps_{t-1}}\left[( \psi(t) (\B{x}_0^\Tgt - \B{x}_0^\Src) + \chi(t) (\hat{\Veps}_{t}^\Tgt - \hat{\Veps}_{t}^\Src) ) \frac{\partial\B{x}_0^{\text{tgt}}}{\partial \theta} \right],
\end{align} 
where $\hat{\Veps}_{t}^{\text{src}} := \Veps_\phi(\B{x}_{t}^{\text{src}}, y^{\text{src}}, t)$ and $\hat{\Veps}_{t}^{\text{tgt}} := \Veps_\phi(\B{x}_{t}^{\text{tgt}}, y^{\text{tgt}}, t)$. We leave a more detailed derivation to the \textbf{supplementary material}.

Matching $\B{z}_t^\Tgt$ with $\B{z}_t^\Src$ ensures that the posteriors of $\B{x}_0^\Tgt$ and $\B{x}_0^\Src$ do not significantly diverge, despite being steered by different prompts, $y^\Tgt$ and $y^\Src$.
This approach is akin to running a generative process with $y^\Tgt$ while remaining near the trajectory made by the posteriors of $\B{x}_0^\Src$. Consequently, PDS enables the sampling of $\B{x}_0^\Tgt$ that aligns with $y^\Tgt$, while also retaining the identity of $\B{x}_0^\Src$. This is achieved through the distillation of the posteriors of $\B{x}_0^\Src$ into the target sampling process.




\subsection{Comparison with SDS~\cite{Poole:2023DreamFusion} and DDS~\cite{Hertz:2023DDS}}

\begin{figure}
\centering
\includegraphics[width=\linewidth]{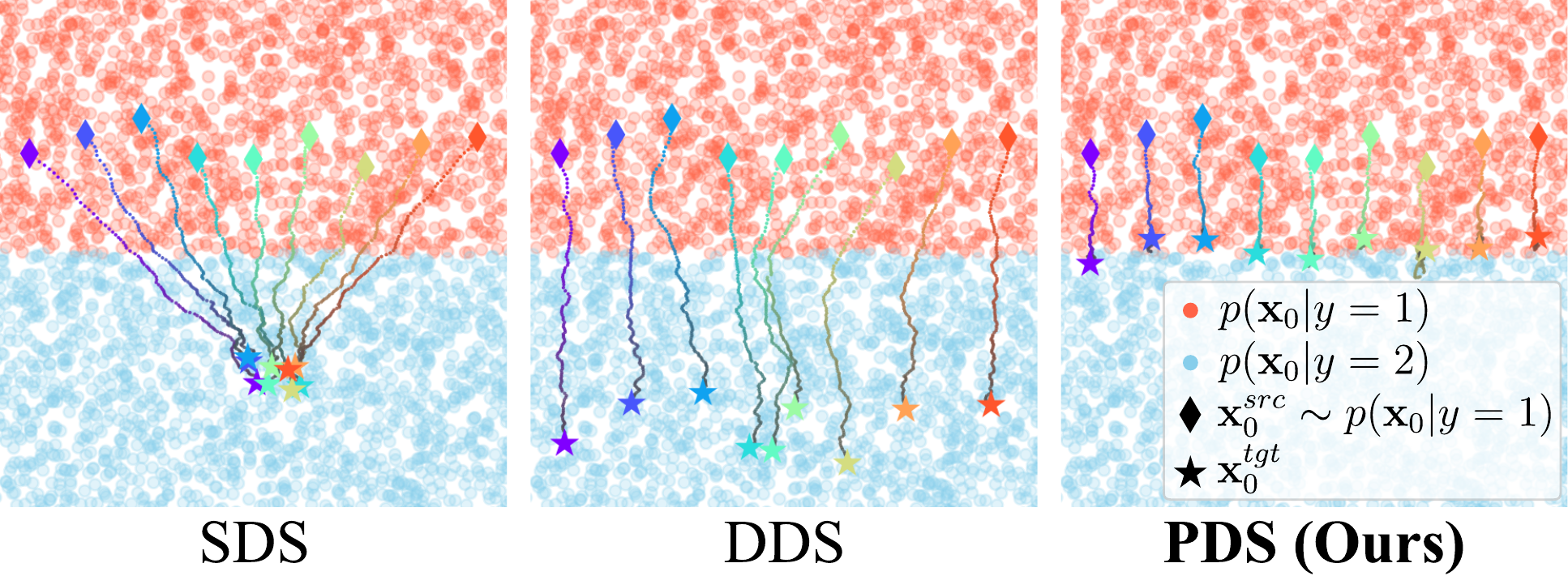}
\caption{\textbf{A visual comparison of the editing process through SDS~\cite{Poole:2023DreamFusion}, DDS~\cite{Hertz:2023DDS} and \Ours{}.} The figure illustrates the trajectories of samples drawn from $p(\B{x}_0 | y=1)$ as they are shifted towards $p(\B{x}_0 | y=2)$. \Ours{} notably moves the samples near the boundary of the two marginals---the optimal endpoint in that it balances the necessary change with the original identity.}
\label{fig:2d_trajectories}
\end{figure}

In Figure~\ref{fig:2d_trajectories}, we visually illustrate the difference among the three optimization methods: SDS~\cite{Poole:2023DreamFusion}, DDS~\cite{Hertz:2023DDS} and \Ours{}. Here, we model a 2D distribution $\B{x}_0 \sim p(\B{x}_0) \in \mathbb{R}^2$ that is separated by two marginals, $p(\B{x}_0 | y=1)$ and $p(\B{x}_0 | y=2)$ which are colored by red and blue, respectively. Then, we train a diffusion model conditioned on the class labels $y$. Using the pre-trained conditional diffusion model, we aim to transition $\B{x}_0^\Tgt$ starting from $\B{x}_0^\Src \sim p(\B{x}_0 | y=1)$ towards the other marginal $p(\B{x}_0 | y=2)$. The trajectories of three optimization methods are plotted in Figure~\ref{fig:2d_trajectories} with their endpoints denoted by stars. As illustrated, SDS and DDS significantly displace the data from the initial position, whereas our method is terminated near the boundary of the two marginals. This is the optimal endpoint for an editing purpose as it indicates proximity to both the starting points and $p(\B{x}_0 | y=2)$, thereby achieving a balance between the necessary change and the original identity.

\subsection{Comparison with Iterative DU}
\label{sec:comparison_with_iterative_du}
\begin{figure}
\centering
\includegraphics[width=\linewidth]{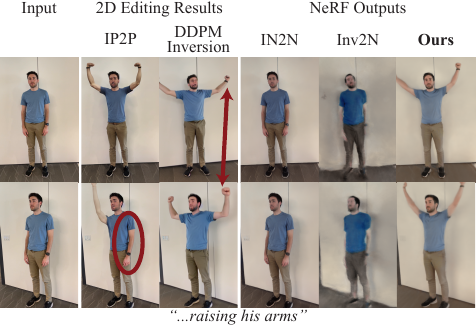}
\caption{\textbf{An example of editing inducing large variations across different views.} The figure shows NeRF editing results of ours and Iterative DU methods, IN2N~\cite{Haque:2023InstructNeRF} and Inv2N, with their corresponding 2D editing results obtained by IP2P~\cite{Brooks:2023InstructPix2Pix} and DDPM Inversion~\cite{Huberman:2023FriendlyInversion}, respectively.
When 2D editing leads to large variations, the Iterative DU methods fail to produce accurate edits in 3D space.}
\label{fig:multiview_inconsistency}
\end{figure}

When a parameterization of images is given as NeRF~\cite{Mildenhall:2020NeRF}, recent works~\cite{Haque:2023InstructNeRF, Mirzae:i2023WatchYourStep} have shown promising NeRF editing results based on a method known as Iterative Dataset Update (Iterative DU). This method bypasses 3D editing by performing the editing process within 2D space. Given an image dataset $\{I^\Src_v\}_{v=1}^N$ used for NeRF~\cite{Mildenhall:2020NeRF} reconstruction with viewpoints $v$, they randomly replace $I^\Src_v$ with its 2D edited version using Instruct-Pix2Pix (IP2P)~\cite{Brooks:2023InstructPix2Pix}. By iteratively updating the input images, they progressively transform the input NeRF scene into an edited version of it.

In contrast to Iterative DU which performs editing in 2D space, our approach directly edits NeRFs~\cite{Mildenhall:2020NeRF} in 3D space.
To visually demonstrate this difference, Figure~\ref{fig:multiview_inconsistency} presents a qualitative comparison of ours and various methods based on Iterative DU. Specifically, we compare ours with Instruct-NeRF2NeRF (IN2N)~\cite{Haque:2023InstructNeRF} which uses IP2P~\cite{Brooks:2023InstructPix2Pix} for 2D editing. Additionally, we include another Iterative-DU-based method, Inversion2NeRF (Inv2N), which employs DDPM inversion~\cite{Huberman:2023FriendlyInversion} for its 2D editing process. Given the prompt \emph{``raising his arms"}, the figure illustrates significant variations in 2D edited images across different views: the man raises either only one arm or both arms, as marked by the red circle. Furthermore, the red arrow highlights the inconsistency in the poses of raising arms across different views. Such notable discrepancies in 2D editing hinder the Iterative DU methods from transferring these edits into 3D space. Particularly noteworthy is the comparison of our method with Inv2N, both of which leverage the stochastic latent for editing. However, while Inv2N confines its editing within 2D space, ours directly updates NeRF parameters in 3D space by reformulating the 2D image editing method~\cite{Huberman:2023FriendlyInversion} into an optimization form. Consequently, as shown in Figure~\ref{fig:multiview_inconsistency} and Figure~\ref{fig:nerf_editing_comparison}, ours is the only one to facilitate complex geometric changes and the addition of objects in 3D scenes. It demonstrates the strength of our method lies in the novel optimization design, which allows for direct 3D editing, not just relying on the editing capabilities of DDPM inversion~\cite{Huberman:2023FriendlyInversion}.

\section{NeRF Editing with \Ours{}}
\label{sec:nerf_editing_with_pds}
As one of the applications of \Ours{}, we present a detailed pipeline for NeRF~\cite{Mildenhall:2020NeRF} editing.
NeRF can be seen as a parameterized rendering function. The rendering process is expressed as $I_v = g(v; \theta)$, where the function takes a specific viewpoint $v$ to render the image $I_v$ at that viewpoint with the rendering parameter $\theta$. Using the publicly available Stable Diffusion~\cite{Rombach:2022StableDiffusion} as our diffusion prior model, we encode the current rendering at viewpoint $v$ to obtain the target latent $\B{x}_{0,v}^\Tgt$: $\B{x}^\Tgt_{0,v} := \mathcal{E}(g(v; \theta))$, where $\mathcal{E}$ is a pre-trained encoder.
Similarly, given the original source images $\{I^\Src_v\}$ used for NeRF~\cite{Mildenhall:2020NeRF} reconstruction, the source latent $\B{x}^\Src_{0,v}$ is also computed by encoding the source image at viewpoint $v$: $\B{x}^\Src_{0,v} := \mathcal{E}(I_v^\Src)$.

For real scenes, there are no given source prompts. Thus, we manually create descriptions for the real scenes, such as \emph{``a photo of a man"} in Figure~\ref{fig:teaser}. For target prompts $y^\Tgt$, we adjust $y^\Src$ by appending a description of a desired attribute---\eg \emph{``...raising his arms"} in Figure~\ref{fig:multiview_inconsistency}---or by substituting an existing word in $y^\Src$ with a new one, such as changing \emph{``deer doll"} to \emph{``unicorn doll"} in the last row of Figure~\ref{fig:nerf_editing_comparison}. Given a pre-fixed set of viewpoints $\{v\}$, we randomly select a viewpoint $v$ to compute $\B{x}_{0,v}^\Src$ and $\B{x}_{0,v}^\Tgt$. The pairs of $(\B{x}_{0,v}^\Src, y^\Src)$ and $(\B{x}_{0,v}^\Tgt, y^\Tgt)$ are fed into the \Ours{} optimization to update $\theta$ in a direction dictated by the target prompt. After the optimization, the updated NeRF parameter $\tilde{\theta}$ renders an edited 3D scene that is aligned with the target prompt: $\tilde{I}_v := g(v; \tilde{\theta})$.

To further improve the final output, we take a refinement stage inspired by DreamBooth3D~\cite{raj2023dreambooth3d}. During iterations of the refinement stage, we randomly select an edited rendering $\tilde{I}_v$ and refine it into a more realistic-looking image using SDEdit~\cite{Meng:2022SDEdit}. The edited NeRF scenes through \Ours{} optimization are then further refined by a reconstruction loss with these repeatedly updated images.

In some cases of source prompts we create, we observe some gap between the ideal text prompt, which would ideally reconstruct the input image through the generative process, and the actual prompt we provide. To alleviate this discrepancy issue, we have found it effective to finetune the Stable Diffusion~\cite{Rombach:2022StableDiffusion} with $\{I^\Src_v\}$ and $y^\Src$ following the DreamBooth~\cite{Ruiz:2023DreamBooth} setup. 


%% file: Sections/5_Experimental_Results.tex
\section{Experiment Results}
\label{sec:experimental_results}
In this section, we conduct editing experiments across two types of parameterized images. Section~\ref{sec:nerf_editing} presents NeRF editing results, comparing our NeRF editing capabilities to the state-of-the-art NeRF editing methods. Furthermore, Section~\ref{sec:svg_editing} shows SVG editing results to compare \Ours{} against other optimization methods, namely SDS~\cite{Poole:2023DreamFusion} and DDS~\cite{Hertz:2023DDS}.
 
\subsection{NeRF Editing}
\label{sec:nerf_editing}
\paragraph{Datasets.}
We use real scenes we capture as well as the scenes from IN2N~\cite{Haque:2023InstructNeRF} and LLFF~\cite{Mildenhall:2019LLFF}. The total number of scenes is \num{13}, and the final number of pairs of source scenes and target text prompts is \num{37} with multiple target prompts for each scene. 

\paragraph{Baselines.}
For extensive comparisons, we evaluate our method against three baselines: Instruct-NeRF2NeRF (IN2N)~\cite{Haque:2023InstructNeRF}, DDS~\cite{Hertz:2023DDS} and Inversion2NeRF (Inv2N).
First, we compare ours with IN2N~\cite{Haque:2023InstructNeRF}, which is a state-of-the-art NeRF editing method with its code publicly available. Additionally, as introduced in Section~\ref{sec:comparison_with_iterative_du}, we conduct a comparison with Inv2N, another method based on Iterative DU, which performs editing within 2D space rather directly in 3D space, but employs DDPM inversion~\cite{Huberman:2023FriendlyInversion} instead of IP2P~\cite{Brooks:2023InstructPix2Pix} for 2D editing.

\paragraph{Results.}
Figure~\ref{fig:nerf_editing_comparison} presents the qualitative comparisons of NeRF editing. Notably, as depicted in rows 1 and 2, our method is the only one that makes large geometric changes in 3D scenes from the input text, folding the man's arms to create natural poses of him reading a book or drinking coffee. In contrast, Iterative-DU-based methods like IN2N~\cite{Haque:2023InstructNeRF} and Inv2N fail to produce the right edits in 3D space. DDS~\cite{Hertz:2023DDS} produces the outputs that completely lose the identity of the input scenes, focusing solely on conforming to the input texts. Rows 3 and 4 of Figure~\ref{fig:nerf_editing_comparison} show the editing scenarios of adding objects in outdoor scenes without specifying local regions, which also leads to large variations. Here, our method successfully adds objects like windmills and hot air balloons in the input scenes, maintaining their background details. On the other hand, the baselines either fail to add the objects in 3D space or produce outputs that significantly deviate from the original scenes. When it comes to appearance change, which induces relatively little variations across different views, both our method and IN2N~\cite{Haque:2023InstructNeRF} effectively produce the desired appearance change in 3D scenes, as shown in the last row of Figure~\ref{fig:nerf_editing_comparison}. However, ours most preserves the original identity of the input scene, such as the object's color, while making appropriate changes. 
\ifpaper
Additional qualitative results are presented through videos on our \href{https://posterior-distillation-sampling.github.io/}{project page}\footnote{\href{https://posterior-distillation-sampling.github.io/}{https://posterior-distillation-sampling.github.io}}.
\else
We provide more qualitative results in the \textbf{supplementary material}.
\fi

To further assess the perceptual quality of the editing results, we conduct a user study compared to the baselines. Following Ritchie~\cite{Ritchie:MTurk}, participants were shown input NeRF scene videos, editing prompts, and edited NeRF scene videos produced by ours and the baselines. They were then asked to choose the most appropriate edited NeRF scene video. As illustrated in Table~\ref{tbl:nerf_editing_results}, our editing results are most preferred over the baselines in human evaluation by a large margin: 49.33\% (Ours) vs. 27.71\% (IN2N~\cite{Haque:2023InstructNeRF}, the second best). See the \textbf{supplementary material} for a more detailed user study setup.

For a quantitative evaluation, we measure CLIP~\cite{Radford:2021CLIP} Score that measures the similarity between edited 2D renderings and target text prompts in CLIP~\cite{Radford:2021CLIP} space. As shown in Table~\ref{tbl:nerf_editing_results}, ours outperforms the baselines quantitatively. This is corroborated by the qualitative results illustrated in Figure~\ref{fig:nerf_editing_comparison}, especially in scenarios of geometric changes or object addition, where the other baselines have difficulty in making the right edits.

\input{Tables/nerf_editing_results}

\subsection{SVG Editing}
\label{sec:svg_editing}

\begin{figure}
\centering
\includegraphics[width=\linewidth]{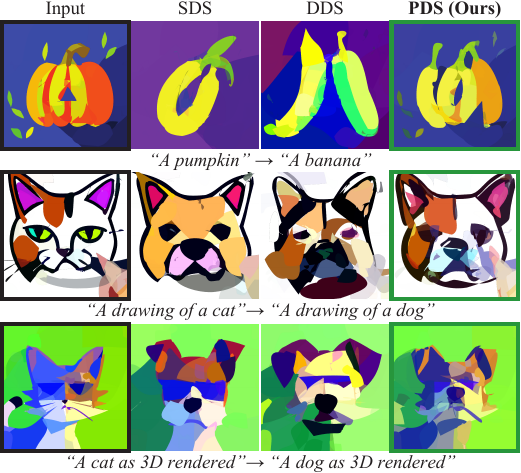}
\caption{\textbf{A qualitative comparison of SVG editing using three different optimization methods: SDS~\cite{Poole:2023DreamFusion}, DDS~\cite{Hertz:2023DDS} and \Ours{}.} \Ours{} makes changes according to input text while most preserving the structural semantics of the input SVGs.}
\vspace{-1\baselineskip}
\label{fig:svg_editing_results}
\end{figure}

\input{Tables/svg_editing_results}

\paragraph{Experimental Setup.}
We use pairs of SVGs and their corresponding text prompts used in VectorFusion~\cite{Jain:2023VectorFusion} as input. By manually creating target text prompts, we conduct experiments with a total of \num{48} pairs of input SVGs and target text prompts. For comparison, we evaluate our method against other optimization methods, SDS~\cite{Poole:2023DreamFusion} and DDS~\cite{Hertz:2023DDS}. To perform editing with SDS, we start with a source SVG as an initial updated SVG and then update it using a target prompt according to the SDS~\cite{Poole:2023DreamFusion} optimization. Following DDS, we use CLIP~\cite{Radford:2021CLIP} score and LPIPS~\cite{Zhang:2018LPIPS} as quantitative metrics.
\vspace{-\baselineskip}
\paragraph{Results.}
Qualitative results of SVG editing are shown in Figure~\ref{fig:svg_editing_results}. It demonstrates that while all the methods effectively change input SVGs according to the target text prompts, ours best preserves the structural semantics of the input SVGs. This is particularly evident in row 2 of Figure~\ref{fig:svg_editing_results}, where ours maintains the overall color pattern of the input SVG.

The trends from the qualitative results are mirrored in our quantitative results. As seen in Table~\ref{tbl:svg_editing_results}, ours significantly surpasses the others in LPIPS~\cite{Zhang:2018LPIPS} by a large margin, which measures the fidelity to the input SVG, while our CLIP score is on par with the others. This demonstrates that our method introduces only minimal necessary changes to meet the described attributes in the target text prompts.

We further provide a user study result of SVG editing in Table~\ref{tbl:svg_editing_results}. We use the same user study setup used in NeRF editing (Section~\ref{sec:nerf_editing}). Consistent with the qualitative and quantitative results, ours are most preferred in human evaluation.

%% file: Tables/nerf_editing_results.tex
\begin{table}[t!]
\centering
\caption{\textbf{A quantitative comparison of NeRF editing between ours and other baselines.} Ours outperforms the baselines quantitatively. \textbf{Bold} indicates the best result for each column.}
{
\scriptsize
\setlength{\tabcolsep}{0.1em}
\begin{tabularx}{\linewidth}{>{\centering}m{2.1cm} | Y | Y }
\toprule
Methods & CLIP~\cite{Radford:2021CLIP} Score $\uparrow$ & \makecell{User Preference\\Rate (\%) $\uparrow$} \\
\midrule
IN2N~\cite{Haque:2023InstructNeRF} & 0.2280 & 27.71 \\
DDS~\cite{Hertz:2023DDS} & 0.2210 & 13.71 \\
Inv2N & 0.2232 & 9.24 \\
\textbf{\Ours{} (Ours)} & \textbf{0.2477} & \textbf{49.33} \\
\bottomrule
\end{tabularx}
}
\label{tbl:nerf_editing_results}
\end{table}

%% file: Tables/svg_editing_results.tex
\begin{table}[t!]
\centering
\caption{\textbf{A quantitative comparison of SVG editing between SDS~\cite{Poole:2023DreamFusion}, DDS~\cite{Hertz:2023DDS} and \Ours{}.} Ours outperforms the others in LPIPS~\cite{Zhang:2018LPIPS} while achieving a CLIP~\cite{Radford:2021CLIP} score that is on par with the others. \textbf{Bold} indicates the best result for each column.}
{
\scriptsize
\setlength{\tabcolsep}{0.2em}
\renewcommand{\arraystretch}{1.0}
\begin{tabularx}{\linewidth}{>{\centering}m{1.6cm}| Y | >{\centering}m{1.6cm} | Y }
\toprule
Methods & CLIP~\cite{Radford:2021CLIP} Score $\uparrow$ & LPIPS~\cite{Zhang:2018LPIPS} $\downarrow$ & \makecell{User Preference\\Rate (\%) $\uparrow$} \\
\midrule 
SDS~\cite{Poole:2023DreamFusion} & \textbf{0.2606} & 0.4855 & 30.83 \\
DDS~\cite{Hertz:2023DDS} & 0.2460 & 0.5982 & 20.24 \\
\textbf{PDS (Ours)} & 0.2504 & \textbf{0.3121} & \textbf{48.94} \\
\bottomrule 
\end{tabularx}
}
\label{tbl:svg_editing_results}
\vspace{-1\baselineskip}
\end{table}

%% file: Sections/6_Conclusion.tex
\vspace{-0.5\baselineskip}
\section{Conclusion}
\label{sec:conclusion}
\vspace{-0.5\baselineskip}
We propose Posterior Distillation Sampling (PDS), an optimization method for parametric image editing. 
PDS matches the stochastic latents of the source and the target to fulfill both conformity to the target text and preservation of the source identity in parameter space. 
We demonstrate the versatility of PDS in parametric image editing through a comparative analysis between ours and other optimization methods and extensive experiments across various parameter spaces.
\vspace{-\baselineskip}
\paragraph{Acknowledgements}
This work was supported by
NRF grant (RS-2023-00209723) and IITP grants (2022-0-
00594, RS-2023-00227592) funded by the Korean government (MSIT), Seoul R\&BD Program (CY230112), and
grants from the DRB-KAIST SketchTheFuture Research Center, Hyundai NGV, KT, NCSOFT, and Samsung Electronics.

%% file: Sections/999_Supplementary.tex
\ifpaper
  \newcommand{\refofpaper}[1]{\unskip}
  \newcommand{\refinpaper}[1]{\unskip}  
\else
  \makeatletter
  \newcommand{\manuallabel}[2]{\def\@currentlabel{#2}\label{#1}}
  \makeatother
  \manuallabel{sec:related_work}{2}
  \manuallabel{sec:preliminaries}{3}
  \manuallabel{sec:nerf_editing}{6.1}
  \manuallabel{sec:svg_editing}{6.2}
  \manuallabel{sec:pds}{4}
  \manuallabel{sec:nerf_editing_with_pds}{5}

  \manuallabel{eq:final_pds_form}{14}
  
  \newcommand{\refofpaper}[1]{of the main paper}
  \newcommand{\refinpaper}[1]{in the main paper}
\fi


\ifpaper
\else

\noindent In this supplementary material,
we first present additional editing results with more diverse representations, including 3D Gaussian Splats (3DGS)~\cite{Kerbl:20233DGS} and 2D images, in Section~\ref{sec:suppl_editing_more_diverse_representations}.
Then, in Section~\ref{sec:suppl_derivation_of_pds}, we provide a more detailed derivation of Posterior Distillation Sampling introduced in Section~\ref{sec:pds}~\refofpaper{}. Following that, Section~\ref{sec:suppl_implementation_details} presents the implementation details of NeRF editing and SVG editing discussed in Sections~\ref{sec:nerf_editing} and~\ref{sec:svg_editing}~\refofpaper{}, respectively. Subsequently, Section~\ref{sec:suppl_details_of_user_study} provides the details of our user study setups. Lastly, the effect of the refinement stage, discussed in Section~\ref{sec:nerf_editing_with_pds}~\refofpaper{}, is detailed in Section~\ref{sec:suppl_refinement_stage}.

\fi

\subsection{Editing 3D Gaussian Splats~\cite{Kerbl:20233DGS} and 2D Images}
\vspace{-\baselineskip}
\label{sec:suppl_editing_more_diverse_representations}
\begin{figure}[h!]
\centering
\captionsetup{type=figure}
\includegraphics[width=\linewidth]{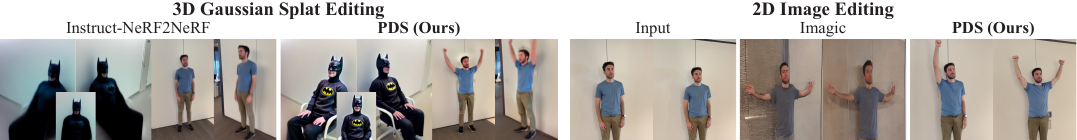}
\caption{\textbf{Editing of more diverse representations, 3D Gaussian Splats~\cite{Kerbl:20233DGS} and 2D images.} PDS consistently outperforms the baselines. The target attributes are \emph{``Batman"} and \emph{``raising the arms."}}
\label{fig:gs_image_editing}
\end{figure}
\vspace{-\baselineskip}
\noindent \Ours{} encompasses various editing scenarios, not confined within a specific parameter space. To further assess the versatility and generalizability of \Ours{} in editing tasks, we include both 3D Gaussian Splat (3DGS)~\cite{Kerbl:20233DGS} editing and 2D image editing. As NeRF editing, Figure~\ref{fig:gs_image_editing} shows that \Ours{} outperforms Instruct-NeRF2NeRF~\cite{Haque:2023InstructNeRF} in 3DGS representation while uniquely realizing geometric changes. In 2D image editing, \Ours{} demonstrates superior performance compared to Imagic~\cite{Kawar:2023Imagic}, which is introduced for 2D image editing using pre-trained 2D diffusion models. \Ours{} edits the input image while preserving other details with high fidelity. On the other hand, Imagic~\cite{Kawar:2023Imagic} leaves artifacts, losing the identity of the source content. 

\subsection{Derivation of Posterior Distillation Sampling}
\label{sec:suppl_derivation_of_pds}
For a comprehensive derivation of Equation~\ref{eq:final_pds_form}~\refinpaper{}, we first remind that the objective function of PDS is expressed as:
\begin{align}
    \mathcal{L}_{\tilde{\B{z}}_t}(\B{x}_0^{\text{tgt}}) &=
    \mathbb{E}\left[
    \Vert \tilde{\B{z}}_{t}^{\text{tgt}} - \tilde{\B{z}}_{t}^{\text{src}} \Vert_2^2\right] \\
    &= \mathbb{E} \left[ \Big\Vert \frac{\B{x}_{t-1}^\Tgt -\Vmu_\phi(\B{x}_t^\Tgt, y^\Tgt; \Veps_\phi)}{\sigma_t} - \frac{\B{x}_{t-1}^\Src - \Vmu_\phi(\B{x}_t^\Src, y^\Src; \Veps_\phi)}{\sigma_t} \Big\Vert_2^2 \right] \\ 
\label{eq:suppl_pds_objective_function}
    &= \mathbb{E} \left[ \frac{1}{\sigma_t^2} \big\Vert (\B{x}_{t-1}^\Tgt - \B{x}_{t-1}^\Src) - \left( \Vmu_\phi(\B{x}_t^\Tgt, y^\Tgt; \Veps_\phi) - \Vmu_\phi(\B{x}_t^\Src, y^\Src; \Veps_\phi)\right)\big\Vert_2^2
    \right].
\end{align}
Given that $\tilde{\B{z}}^\Src_t$ and $\tilde{\B{z}}^\Tgt_t$ share the same noises $\Veps_{t-1}$ and $\Veps_{t}$ for their respective $\B{x}_{t-1}$ and $\B{x}_t$, the difference between $\B{x}_{t-1}^\Tgt$ and $\B{x}_{t-1}^\Src$ results in a constant multiple of the difference between $\B{x}_0^\Tgt$ and $\B{x}_0^\Src$:
\begin{align}
\label{eq:x_tm1_residual}
    \B{x}_{t-1}^\Tgt - \B{x}_{t-1}^\Src &= \sqrt{\bar\alpha_{t-1}} ( \B{x}_0^\Tgt - \B{x}_0^\Src).
\end{align}
Following our notation $\hat{\Veps}_t^\Src := \Veps_\phi(\B{x}_t^\Src, y^\Src, t)$ and $\hat{\Veps}_t^\Tgt := \Veps_\phi(\B{x}_t^\Tgt, y^\Tgt, t)$ introduced in Section~\ref{sec:pds}~\refofpaper{}, the difference between the approximated posterior means is also expressed as follows:
\begin{align}
\label{eq:mean_residual}
    \Vmu_\phi(\B{x}_t^\Tgt, y^\Tgt; \Veps_\phi) - \Vmu_\phi(\B{x}_t^\Src, y^\Src, \Veps_\phi) &= (\gamma_t + \delta_t \sqrt{\bar\alpha_t})(\B{x}_0^\Tgt - \B{x}_0^\Src) -\gamma_t \sqrt{\frac{1}{\bar\alpha_t} - 1} (\hat{\Veps}_t^\Tgt - \hat{\Veps}_t^\Src),
\end{align}
where  $\Vmu_\phi(\B{x}_t, y; \Veps_\phi)$ can be expanded as shown in the following equation:
\begin{align}
    \Vmu_\phi(\B{x}_t, y; \Veps_\phi) &= \gamma_t \tilde{\B{x}}_0(\B{x}_t, y; \Veps_\phi) + \delta_t \B{x}_t \\
    &= \gamma_t \left( \frac{1}{\sqrt{\bar\alpha_t}}(\B{x}_t - \sqrt{1 - \bar\alpha_t} \Veps_\phi(\B{x}_t, y, t) \right) + \delta_t \B{x}_t \\
    &= (\frac{\gamma_t}{\sqrt{\bar{\alpha}}_t}+\delta_t)\B{x}_t - \gamma_t \sqrt{\frac{1}{\bar{\alpha}_t} - 1}\Veps_\phi(\B{x}_t,y,t) \\ 
    &= (\gamma_t+\delta_t\sqrt{\bar\alpha_t})\B{x}_0 + \sqrt{\frac{1}{\bar\alpha_t}-1} (\gamma_t+\delta_t\sqrt{\bar\alpha_t}) \Veps_t - \gamma_t \sqrt{\frac{1}{\bar{\alpha}_t} - 1}\Veps_\phi(\B{x}_t,y,t).
\end{align}
Incorporating Equation~\ref{eq:x_tm1_residual} and Equation~\ref{eq:mean_residual} into Equation~\ref{eq:suppl_pds_objective_function}, we can reformulate the objective function of PDS as follows:
\begin{align}
\label{eq:expanded_objective_function}
    \mathcal{L}_{\tilde{\B{z}}_t}(\B{x}_0^{\text{tgt}}) = \mathbb{E} \biggl[ \frac{1}{\sigma_t^2}&
    \big\Vert (\sqrt{\bar\alpha_{t-1}} - \gamma_t - \delta_t \sqrt{\bar\alpha_t}) (\B{x}_0^\Tgt - \B{x}_0^\Src) + \gamma_t \sqrt{\frac{1}{\bar\alpha_t} - 1} (\hat{\Veps}^\Tgt_t - \hat{\Veps}^\Src_t) \big\Vert_2^2
    \biggl] \\
    = \mathbb{E} \biggl[ \frac{1}{\sigma_t^2}& \biggl( 
    (\sqrt{\bar\alpha_{t-1}} - \gamma_t - \delta_t \sqrt{\bar\alpha_t})^2 (\B{x}_0^\Tgt - \B{x}_0^\Src)^2 \\ 
    &+ 2(\sqrt{\bar\alpha_{t-1}} - \gamma_t - \delta_t \sqrt{\bar\alpha_t})\gamma_t \sqrt{\frac{1}{\bar\alpha_t} - 1}(\B{x}_0^\Tgt - \B{x}_0^\Src)(\hat{\Veps}^\Tgt_t - \hat{\Veps}^\Src_t) \nonumber \\
    &+ \gamma_t^2 (\frac{1}{\bar\alpha_t} - 1)(\hat{\Veps}^\Tgt_t - \hat{\Veps}^\Src_t)^2
    \biggl) \nonumber \biggl].
\end{align}
By taking the gradient of $\mathcal{L}_{\tilde{\B{z}}_t}$ with respect to $\theta$ while ignoring the U-Net jacobian term, $\frac{\partial \hat{\Veps}_\phi^\Tgt}{\partial \B{x}_0^\Tgt}=\mathbf{I}$, one can obtain PDS as follows: 
\begin{align}
    \nabla_\theta \mathcal{L}_{\text{PDS}} &= \frac{\partial \mathcal{L}_{\tilde{\B{z}}_t}(\B{x}_0^\Tgt)}{\partial \B{x}_0^\Tgt} \cdot \frac{\partial \B{x}_0^\Tgt}{\partial \theta} \\
    &= 
    \mathbb{E} \left[ \frac{2}{\sigma_t^2}\left((\sqrt{\bar\alpha_{t-1}} - \gamma_t -\delta_t \sqrt{\bar\alpha_t})^2 (\B{x}_0^\Tgt - \B{x}_0^\Src)
    +(\sqrt{\bar\alpha_{t-1}} - \gamma_t -\delta_t \sqrt{\bar\alpha_t})\gamma_t\sqrt{\frac{1}{\bar\alpha_t} - 1} (\hat{\Veps}^\Tgt_t - \hat{\Veps}^\Src_t) \right) \frac{\partial \B{x}_0^\Tgt}{\partial \theta}
    \right].
\end{align}
Thus, the coefficients $\psi(t)$ and $\chi(t)$ in Equation~\ref{eq:final_pds_form}~\refofpaper{} are as follows:
\begin{align}
    \psi(t) &= \frac{2(\sqrt{\bar\alpha_{t-1}} - \gamma_t -\delta_t \sqrt{\bar\alpha_t})^2}{\sigma_t^2}, \\
    \chi(t) &= \frac{2(\sqrt{\bar\alpha_{t-1}} - \gamma_t -\delta_t \sqrt{\bar\alpha_t})}{\sigma_t^2}\gamma_t\sqrt{\frac{1}{\bar\alpha_t} - 1}.
\end{align}
In practice, we sample non-consecutive timesteps for $t-1$ and $t$ as in DDIM~\cite{Song:2021DDIM} since the coefficients become $0$ when they are consecutive. Given a sequence of non-consecutive timesteps $[\tau_i]_{i=1}^S$, a more generalized form of PDS is represented as follows:
\begin{align}
    \nabla_\theta \mathcal{L}_{\text{PDS}} = \mathbb{E}_{i, \Veps_{\tau_i}, \Veps_{\tau_{i-1}}} \left[ 
    \psi(i) (\B{x}_0^\Tgt - \B{x}_0^\Src) + \chi(i) (\hat{\Veps}_{\tau_i}^\Tgt - \hat{\Veps}_{\tau_i}^\Src) \frac{\partial \B{x}_0^\Tgt}{\partial \theta}
    \right],
\end{align}
where
\begin{align}
    \psi(i) &= \frac{2(\sqrt{\bar\alpha_{\tau_{i-1}}} - \gamma_{\tau_i} -\delta_{\tau_i} \sqrt{\bar\alpha_{\tau_i}})^2}{\sigma_{\tau_i}^2}, \\
    \chi(i) &= \frac{2(\sqrt{\bar\alpha_{\tau_{i-1}}} - \gamma_{\tau_i} -\delta_{\tau_i} \sqrt{\bar\alpha_{\tau_i}})}{\sigma_{\tau_i}^2}\gamma_{\tau_i}\sqrt{\frac{1}{\bar\alpha_{\tau_i}} - 1}.
\end{align}
For more details on timestep sampling, refer to the implementation details in the next section.

\subsection{Implementation Details}
\label{sec:suppl_implementation_details}
In this section, we provide the implementation details of NeRF and SVG editing presented in Section~\ref{sec:nerf_editing} and Section~\ref{sec:svg_editing}~\refofpaper{}, respectively.

\paragraph{NeRF Editing.}
 We run the \Ours{} optimization for \num{30000} iterations with classifier-free guidance~\cite{Ho:2021CFG} weights within $[30, 100]$ depending on the complexity of editing. As detailed in Section~\ref{sec:suppl_derivation_of_pds}, we sample non-consecutive timesteps $\tau_{i-1}$ and $\tau_i$ since the coefficients $\psi(\cdot)$ and $\chi(\cdot)$ become zero when the sampled timesteps are consecutive. For this, we define non-consecutive timesteps $[\tau_i]_{i=1}^S$, which is a subset sequence of the total forward process timesteps of the diffusion model, $[1, ..., T]$. Specifically, we select these timesteps such that $\tau_i = \lfloor{2i}\rfloor$, resulting in a subset sequence length of $S=500$ out of the total $T=1000$ timesteps. We then randomly sample the index $i$ within a ratio range of $[0.02, 0.98]$,~\ie{} $i\sim \mathcal{U}(10, 490)$.
 
During the refinement stage, we randomly choose and replace $\tilde{I}_v$ every \num{10} iterations, over total \num{15000} iterations. We denote a SDEdit~\cite{Meng:2022SDEdit} operator by $\mathcal{S}(\B{x}_0;t_0,\Veps_\phi)$ which samples $\B{x}_{t_0} \sim \mathcal{N}(\sqrt{\bar\alpha_{t_0}}\B{x}_0, (1-\bar\alpha_{t_0}) \B{I})$ then starts denoising it from $t_0$ using $\Veps_\phi$. For the denoising process, we randomly sample $t_0$ within a ratio range of $[0, 0.2]$ out of total denoising steps $N=20$.

\paragraph{SVG Editing.}
Across all optimizations, SDS~\cite{Poole:2023DreamFusion}, DDS~\cite{Hertz:2023DDS}, and our proposed PDS, we apply the same classifier-free guidance weight of 100.
For SDS~\cite{Poole:2023DreamFusion}, we sample $t$ within a ratio range of $[0.05, 0.95]$ following VectorFusion~\cite{Jain:2023VectorFusion}. For DDS~\cite{Hertz:2023DDS}, we follow its original setup, sampling $t$ within $[0.02, 0.98]$. For PDS, we sample $i$ out of a ratio range of $[0.1, 0.98]$.


\begin{figure}
\centering
\includegraphics[width=1\linewidth]{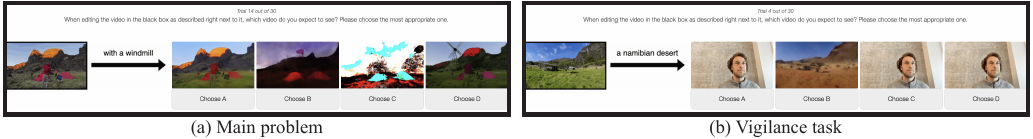}
\caption{\textbf{NeRF editing user study screenshots.} The participants are presented with NeRF scene videos and editing prompts, and are asked to answer the following question: \texttt{When editing the video in the black box as described right next to it, which video do you expect to see? Please choose the most appropriate one.}}
\label{fig:user_study_nerf}
\vspace{-0.5\baselineskip}
\end{figure}

\begin{figure}
\centering
\includegraphics[width=1\linewidth]{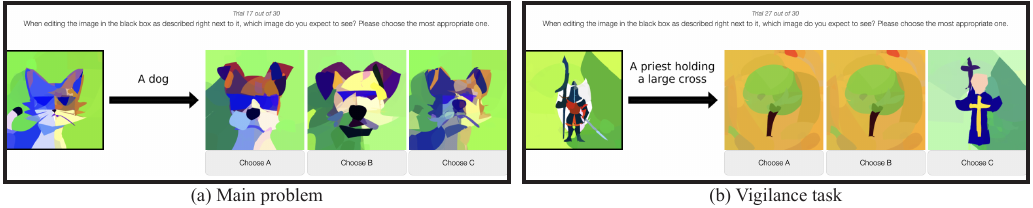}
\caption{\textbf{SVG editing user study screenshots.} Given SVG images and editing prompts, the participants are asked to answer the following question: \texttt{When editing the image in the black box as described right next to it, which image do you expect to see? Please choose the most appropriate one.}}
\label{fig:user_study_svg}
\vspace{-0.5\baselineskip}
\end{figure}

\begin{figure}
\centering
\captionsetup{type=figure}
\includegraphics[width=\textwidth]{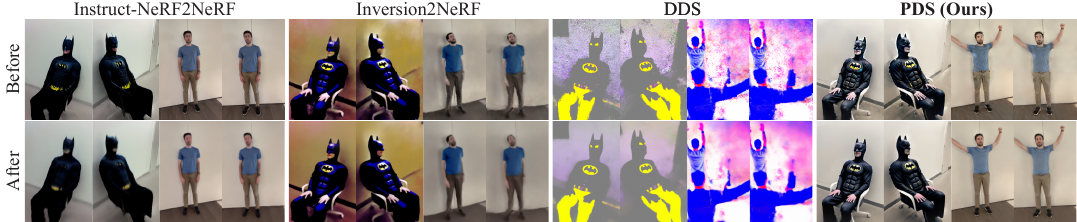}
\vspace{-20pt}
\caption{\textbf{The effect of the refinement stage.} The overall editing outcomes are determined before the refinement stage, whereas the refinement stage plays the role of removing artifacts.
The target attributes are \emph{``Batman"} and \emph{``raising the arms."}
}
\label{fig:refinement_stage}
\vspace{-0.5\baselineskip}
\end{figure}


\subsection{Details of User Studies}
\label{sec:suppl_details_of_user_study}
We conduct user studies for the human evaluation of NeRF and SVG editing through Amazon's Mechanical Turk. We collected survey responses only from those participants who passed our vigilance tasks. 
To design our vigilance tasks, we create examples where, except for the correct answer choice, all other choices are replaced with ones from different scenes or unrelated SVG examples. Screenshots of our NeRF and SVG editing user studies, including examples of vigilance tasks, are displayed in Figure~\ref{fig:user_study_nerf} and Figure~\ref{fig:user_study_svg}, respectively. In the NeRF and SVG editing user studies, we received 42 and 17 valid responses, respectively.

\subsection{Effect of the Refinement Stage}
\label{sec:suppl_refinement_stage}
Figure~\ref{fig:refinement_stage} illustrates an ablation study of the refinement stage across various editing methods. As depicted, the desired complex edits --- making the man raise his arms --- are achieved solely through the optimization of \Ours{}. The overall editing outcomes are realized before the refinement stage, and the refinement stage further enhances the fidelity of the outputs.